\documentclass[runningheads]{llncs}
\usepackage{graphicx}
\usepackage{utils}

\begin{document}
\title{Controlling Model Complexity in Probabilistic Model-Based Dynamic Optimization of Neural Network Structures\thanks{Accepted as a conference paper at the 28th International Conference on Artificial Neural Networks (ICANN 2019). The final authenticated publication will be available in the Springer Lecture Notes in Computer Science (LNCS).}}
\titlerunning{Controlling Model Complexity in Dynamic Optimization of NN Structures}
\author{
Shota Saito\inst{1,2} \and
Shinichi Shirakawa \inst{1}
}

\authorrunning{S. Shota \andname~S. Shirakawa}

\institute{
Yokohama National University, Kanagawa, Japan \\ \email{ \{saito-shota-bt, shirakawa-shinichi-bg\} @ \{ynu, ynu.ac\}.jp } \and 
SkillUp AI Co., Ltd., Tokyo, Japan \\
\email{s\_saito@skillupai.com}
}

\maketitle              % typeset the header of the contribution
\begin{abstract}
A method of simultaneously optimizing both the structure of neural networks and the connection weights in a single training loop can reduce the enormous computational cost of neural architecture search. We focus on the probabilistic model-based dynamic neural network structure optimization that considers the probability distribution of structure parameters and simultaneously optimizes both the distribution parameters and connection weights based on gradient methods. Since the existing algorithm searches for the structures that only minimize the training loss, this method might find overly complicated structures. In this paper, we propose the introduction of a penalty term to control the model complexity of obtained structures. We formulate a penalty term using the number of weights or units and derive its analytical natural gradient. The proposed method minimizes the objective function injected the penalty term based on the stochastic gradient descent. We apply the proposed method in the unit selection of a fully-connected neural network and the connection selection of a convolutional neural network. The experimental results show that the proposed method can control model complexity while maintaining performance.
\end{abstract}

\section{Introduction}
Deep neural networks (DNNs) are making remarkable progress in a variety of tasks, such as image recognition and machine translation. While various neural network structures have been developed to improve predictive performance, the selection or design of neural network structures remains the user's task. In general, tuning a neural network structure improves model performance. It is, however, tedious, because the user must design an appropriate structure for the target task through trial-and-error.

To automate neural network structure design processes, methods called neural architecture search have been developed. A popular approach is to regard the structure parameters (e.g., the numbers of layers and units, the type of layers, and the connectivity) as the hyperparameters and optimized them through black-box optimization methods, such as the evolutionary algorithms \cite{Real2017,Suganuma2017} and Bayesian optimization \cite{Kandasamy2018}. Another approach trains the neural network that generates the network architecture using policy gradient-based reinforcement learning methods \cite{Zoph2017}. However, these approaches require huge computational resources; several works conducted the experiments using more than 100 GPUs \cite{Real2017,Zoph2017}, as the evaluation of a candidate structure requires model training and takes several hours in the case of DNNs.

To solve the computational bottleneck, alternative methods that simultaneously optimize both the structure and the connection weights in a single training loop have been proposed \cite{Liu2019,Pham2018,Shirakawa2018}. These methods are promising because they can find structures with better prediction performance using only one GPU. In this paper, we employ the \emph{dynamic structure optimization} framework introduced in \cite{Shirakawa2018} as the baseline algorithm. This framework considers the probability distribution of structure parameters and simultaneously optimizes both the distribution parameters and weights based on gradient methods.

The above-mentioned methods concentrate on finding neural network structures demonstrating high prediction performance; that is, they search for a structure that minimizes validation or training error. The structures found based on such criteria might become resource-hungry. To deploy such neural networks using limited computing resources, such as mobile devices, a compact yet high-performing structure is required. Several studies introduced the model complexity-based objective function, such as the total number of weights and/or FLOPs. Tan et al. \cite{Tan2018} introduced latency (delay time with respect to (\wrt) data transfer) to the objective function as the penalty in the policy gradient-based neural architecture search and searched for a platform-aware structure. Additionally, multi-objective optimization methods have been applied to obtain the structures over a trade-off curve of the performance and model complexity \cite{Dong2018,Elsken2018}. However, such methods require greater computational resource, as existing methods are based on hyperparameter optimization. 

For the purpose of obtaining compact structures, regularization-based connection pruning methods have been investigated. Han et al. \cite{Han2015} used L2 norm regularization of weights and iterate the weight coefficient-based pruning and retraining. This method obtained a simpler structure with the same performance as the original structure. Liu et al. \cite{Liu2017} proposed channel-wise pruning for use with convolutional neural networks (CNNs) with the addition of new weights for each channel; the weights are penalized through L1 norm regularization. In general, regularization-based pruning methods impose a penalty on the weight values. It is, therefore, difficult to directly use aspect of the network size, such as the number of weight parameters or units, as the penalty.

In this paper, we introduce a penalty term for controlling model complexity in the dynamic structure optimization method \cite{Shirakawa2018}. In accordance with the literature \cite{Shirakawa2018}, we assume the binary vector as the structure parameters that can use to represent the network structure, such as the selection of units or connections between layers. Further, we consider the multivariate Bernoulli distribution and formulate the objective function to be minimized as the expectation of loss function under the distribution. Then the penalty term $\wrt$ the number of weights or units is incorporated into the loss to control the model complexity. To investigate the effects of the proposed penalty term, we apply this method in the unit selection of a fully-connected neural network. The experimental result shows that the proposed method can control the model complexity and preferentially remove insignificant units and connections  to maintain the performance.

\section{The Baseline Algorithm}
\label{sec::related}
We will now briefly explain the dynamic structure optimization framework proposed in \cite{Shirakawa2018}. Neural networks are modeled as $\phi(W, M)$ by two types of parameters: the vector of connection weights $W$ and the structure parameter $M$. The structure parameter $M \in \mathcal{M}$ determines $d$ hyperparameters, such as the connectivity of each layer or the existence of units. Let us consider that the structure parameter $M$ is sampled from the probabilistic distribution $p(M \mid \theta)$, which is parameterized by a vector $\theta \in \Theta$ as a distribution parameter. We denote the loss to be minimized as $\mathcal{L}(W, M) = \int l(z, W, M) p(z) \mathrm{d}z$, where $l(z, W, M)$ and $p(z)$ indicate the loss of a datum $z$ and the probability distribution of $z$, respectively.

Instead of directly optimizing $\mathcal{L}(W, M)$, the \emph{stochastic relaxation} of $M$ is considered; that is, the following expected loss under $p(M \mid \theta)$ is minimized:
\begin{align}
\label{eq::exp_loss_g}
	\mathcal{G}(W, \theta) = \int \mathcal{L}(W, M) p(M \mid \theta) \mathrm{d}M \enspace,
\end{align}
where $\mathrm{d}M$ is a reference measure on $\mathcal{M}$. To optimize $W$ and $\theta$, we use the following vanilla (Euclidian) gradient \wrt\ $W$ and the natural gradient \wrt\ $\theta$:
\begin{align}
	\label{eq::exp_grad_w}
	\nabla_{W} \mathcal{G}(W, \theta) &= \int \nabla_{W} \mathcal{L}(W, M) p(M \mid \theta) \mathrm{d}M \enspace, \\
	\label{eq::exp_grad_t}
	\tilde{\nabla}_{\theta} \mathcal{G}(W, \theta) &= \int \mathcal{L}(W, M) \tilde{\nabla}_{\theta} \ln p(M \mid \theta)  p(M \mid \theta) \mathrm{d}M \enspace,
\end{align}
where $\tilde{\nabla}_{\theta} = F(\theta)^{-1} \nabla_{\theta}$ is the so-called natural gradient \cite{Amari1998} and $F(\theta)$ is the Fisher information matrix of $p(M \mid \theta)$. Optimizing $\theta$ using \eqref{eq::exp_grad_t} works the same way as information geometric optimization (IGO) \cite{Ollivier2017}, which is a unified framework for probabilistic model-based evolutionary algorithms. Different from the IGO, Shirakawa et al. \cite{Shirakawa2018} proposed the simultaneously updating of $W$ and $\theta$ with the gradient directions using \eqref{eq::exp_grad_w} and \eqref{eq::exp_grad_t}, and produced dynamic structure optimization. In practice, the gradients \eqref{eq::exp_grad_w} and \eqref{eq::exp_grad_t} are approximated by Monte-Carlo methods using the mini-batch data samples and the $\lambda$ structure parameters sampled from $p(M \mid \theta)$.

\section{Introducing Penalty Term in Dynamic Structure Optimization}
\label{sec::proposed}
In this section, we introduce a penalty term to dynamic structure optimization to control model complexity. We focus on the case that the structure parameter can be treated as a binary vector, as was done in \cite{Shirakawa2018}.

\subsubsection{Representation of Structure Parameter}
We denote neural networks as $\phi(W, M)$ modeled by the two parameters: $W$ is the weight vector, and $M = (m_1, \dots, m_d)^\T \in \mathcal{M} = \{ 0, 1 \}^d$ is a $d$-dimensional binary vector that determines neural network structures. We consider the multivariate Bernoulli distribution defined by $p(M \mid \theta) = \prod^{d}_{i=1} \theta_{i}^{m_{i}} (1 - \theta_{i})^{1 - m_{i}}$
to be the probability distribution for the random variable $M$, where $\theta = (\theta_{1}, \dots, \theta_{d})^{\T}$, $\theta_{i} \in [ 0, 1 ]$ refers to the parameters of the Bernoulli distribution. For instance, in the connection selection, the parameter $m_i$ determines whether or not the $i$-th connection appears, and $\theta_{i}$ corresponds to the probability that $m_i$ becomes one. 

\subsubsection{Incorporating Penalty Term into Objective Function}
We denote the original loss function of neural network models by $\mathcal{L}(W, M)$, which depends on $W$ and $M$. To penalize the complicated structure, we introduce the penalty term $\mathcal{R}(M)$, which depends on $M$, and obtain the objective function represented by $\mathcal{L}(W, M) + \epsilon \mathcal{R}(M)$, where $\epsilon$ is a penalty coefficient. In this paper, we particularly focus on the case that the penalty term can be represented by the weighted sum of $m_i, \enspace i=1, \dots d$, namely $\mathcal{R}(M) = \sum_{i=1}^d c_i m_i$
where $c_i$ indicates the coefficient representing the model complexity that corresponds to the $i$-th bit. Here, we assume that the model complexity increases if the bit $m_i$ becomes one. This is a reasonable assumption because the binary vector is usually used to determine the existences of connections, layers, and units. Therefore, we can consider that the model complexity increases as the number of `1' bits increases.

As both the original loss and the penalty term are not differentiable $\wrt$ $M$, we employ stochastic relaxation by taking the expectation of the objective function. The expected objective function incorporated with the penalty term under the Bernoulli distribution $p(M \mid \theta)$ is given by
\begin{align}
\label{eq::obj_finc}
    \mathcal{G}(W, \theta)
    &= \sum_{M \in \mathcal{M}} \mathcal{L}(W, M) p(M \mid \theta) + \epsilon \sum^{d}_{i=1} c_{i} \theta_{i} \enspace.
\end{align} 
When $\epsilon=0$, the minimization of $\mathcal{G}(W, \theta)$ recovers the same algorithm with \cite{Shirakawa2018}.

\subsubsection{Gradients for Weights and Distribution Parameters}
To simultaneously optimize $W$ and $\theta$, we derive the gradients of $\mathcal{G}(W, \theta)$ $\wrt$ $W$ and $\theta$. The vanilla gradient $\wrt$ $W$ is given by $\nabla_{W} \mathcal{G}(W, \theta) = \sum_{M \in \mathcal{M}} \nabla_{W} \mathcal{L}(W, M)$ since the penalty term $\mathcal{R}(M)$ does not depend on $W$.
Note that the gradient $\nabla_{W} \mathcal{L}(W, M)$ can be computed through back-propagation.

Regarding the distribution parameters $\theta$, we derive the natural gradient \cite{Amari1998}, defined by the product of the inverse of Fisher information matrix and the vanilla gradient, that is the steepest direction of $\theta$ when the KL-divergence is considered as the pseudo distance of $\theta$.
Since we are considering the Bernoulli distribution$F(\theta)^{-1}$ can be obtained analytically by $F(\theta)^{-1} = \mathrm{diag} ( \theta (1 - \theta))$, where the product of vectors indicates the element-wise product. We then obtain the analytical natural gradients of the log-likelihood and the penalty term as $\tilde{\nabla}_{\theta} \ln p(M \mid \theta) = M - \theta$ and $\tilde{\nabla}_{\theta} \sum^{d}_{i=1} c_{i} \theta_{i} = c \theta (1 - \theta)$, respectively,
where $c = (c_{1}, \dots, c_{d})^{\T}$ is the vector representation of the model complexity coefficients and $\tilde{\nabla}_{\theta} = F(\theta)^{-1} \nabla_{\theta}$ indicates the natural gradient operator. As a result, we obtain the following gradient:
\begin{align}
	\label{eq::natural_grad_ber}
    \tilde{\nabla}_{\theta} \mathcal{G}(W, \theta) = \sum_{M \in \mathcal{M}} & \mathcal{L}(W, M) (M - \theta) + \epsilon c \theta (1 - \theta) \enspace.
\end{align}

\subsubsection{Gradient Approximation}
In practice, the analytical gradients are approximated by Monte-Carlo method using $\lambda$ samples $\{ M_1, \dots, M_\lambda \}$ drawn from $p(M \mid \theta)$. Moreover, the loss $\mathcal{L}(W, M_{i})$ is also approximated using $\bar{N}$ mini-batch samples $\mathcal{Z} = \{ z_{1}, \dots, z_{\bar{N}} \}$. Referring to \cite{Shirakawa2018}, we use the same mini-batch between different $M_{i}$ to obtain an accurate ranking of losses. The approximated loss is given by $\bar{\mathcal{L}}(W, M_{i}) = \bar{N}^{-1} \sum_{z \in \mathcal{Z}} l(z, W, M_{i})$, 
where $l(z, W, M_{i})$ represents the loss of a datum. The gradient for $W$ is estimated by Monte-Carlo method using $\lambda$ samples:
\begin{align}
\label{eq:est_gradient_W_g}
    \nabla_{W} \mathcal{G}(W, \theta) \approx \frac{1}{\lambda} \sum_{i=1}^{\lambda} \nabla_{W} \bar{\mathcal{L}}(W, M_{i}) \enspace.
\end{align}
We can update $W$ using any stochastic gradient descent (SGD) method with \eqref{eq:est_gradient_W_g}.

To update the distribution parameters $\theta$, we transform the loss value $\bar{\mathcal{L}}(W, M_{i})$ into the ranking-based utility $u_{i}$ as was done in \cite{Shirakawa2018}: $u_i = 1$ if $\bar{\mathcal{L}}(W, M_i)$ is in top $\lceil\lambda/4\rceil$, $u_i=-1$ if it is in bottom $\lceil\lambda/4\rceil$, and $u_i=0$ otherwise.
The ranking-based utility transformation makes the algorithm invariant to the order preserving transformation of $\mathcal{L}$. We note that this utility function transforms the original minimization problem into a maximization problem. With this utility transformation, the approximation of \eqref{eq::natural_grad_ber} is given by $\tilde{\nabla}_{\theta} \mathcal{G}(W, \theta) \approx \frac{1}{\lambda} \sum_{i=1}^{\lambda} u_{i} (M_{i} - \theta) - \epsilon c \theta (1 - \theta)$.
As a result, the update rule for $\theta$ at the $t$-th iteration is given by
\begin{align}
    \label{eq:update_theta}
    \theta^{(t+1)} = \theta^{(t)} + \eta_{\theta} \bigg ( &\sum_{i=1}^{\lambda} \frac{u_{i}}{\lambda} (M - \theta^{(t)}) - \epsilon c \theta^{(t)} (1 - \theta^{(t)}) \bigg ) \enspace,
\end{align}
where $\eta_{\theta}$ is the learning rate for $\theta$. If we use the binary vector to select input units and set $c = (1, \dots, 1)^\T$, the algorithm works as feature selection \cite{Saito2018}. The method introduced in this paper targets model complexity control and can be applied in cases where each bit corresponds to a different number of weights by introducing the model complexity coefficient $c$.
The optimization procedure of the proposed method is shown in Algorithm \ref{alg:update_procedure}.

\subsubsection{Prediction for Test Data}
As was proposed in \cite{Shirakawa2018}, there are two options for predicting new data using optimized $\theta$ and $W$. In the first method, the binary vectors are sampled from $p(M \mid \theta)$, and the prediction results are averaged. This stochastic prediction method will produce an accurate prediction, but it is not a desirable to obtain a compact structure. The second way is to deterministically select the binary vector as $M^* = \argmax_{M} p(M \mid \theta)$ such that $m_i = 1$ if $\theta_{i} \geq 0.5$; otherwise, $m_i = 0$. In our experiment, we use the second option, deterministic prediction, and report our results.

\begin{algorithm}[tb]
{\small
    \caption{The training procedure of the proposed method.}
    \label{alg:update_procedure}
    \setstretch{0.9}
    \DontPrintSemicolon
    \KwIn{Training data $\mathcal{D}$ and hyperparameters $\{ \lambda,\; \eta_{\theta},\; \epsilon' \}$}
    \KwOut{Optimized parameters of $W$ and $\theta$}
    \Begin{
        Initialize the weight and distribution parameters as $W^{(0)}$ and $\theta^{(0)}$\;
        $t \leftarrow 0$ \;
        \While{not stopping criterion is satisfied}{
            Get $\bar{N}$ mini-batch samples from $\mathcal{D}$\;
            Sample $M_{1}, \dots, M_{\lambda}$ from $p(M \mid \theta^{(t)})$\;
            Compute the losses $\bar{\mathcal{L}}(W, M_i)$ for $i=1, \dots, \lambda$\;
            Update the distribution parameters to $\theta^{(t+1)}$ by \eqref{eq:update_theta}\;
            Restrict the range of $\theta^{(t+1)}$\;
            Update the weight parameters to $W^{(t+1)}$ using \eqref{eq:est_gradient_W_g} by any SGD\;
            $t \leftarrow t+1$\;
        }
    }
}
\end{algorithm}

\subsubsection{Implementation Remark}
We restrict the range of $\theta$ within $[1/d, 1 - 1/d]$ to retain the possibility of generating any binary vector. To be precise, if the updated $\theta$ through \eqref{eq:update_theta} falls outside this range, the values of $\theta$ are set at the boundary value. 
In addition, the coefficient of $\epsilon$ is normalized as $\epsilon = \epsilon' / \max ( c )$.
The natural gradient corresponding to $\mathcal{L}$ is bounded within the range of $[-1, 1]^{d}$ due to the utility transformation. In the above normalization, the one corresponding to the penalty term is bounded within $[0, \epsilon' / 4]^{d}$. Therefore, both the gradients are at approximately the same scale regardless of their encoding scheme (i.e., the usage of $M$).

\section{Experiment and Result}
\label{sec::exp_and_res}
We apply the proposed method to the two neural network structure optimization problems with image classification datasets: unit selection of fully-connected neural networks and connection selection of DenseNet \cite{Huang2017}. All algorithms are implemented using Chainer framework \cite{Tokui2015} (version 4.5.0) and CuPy backend \cite{Okuta2017} (version 4.5.0) and run on a single NVIDIA GTX 1070 GPU in the experiment of unit selection and on a single NVIDIA GTX 1080Ti GPU in the experiment of connection selection. In both experiments, weights are optimized using the SGD with Nesterov's momentum. The coefficient of the momentum and the weight decay are set to 0.9 and $10^{-4}$, respectively. Based on \cite{Huang2017,Shirakawa2018}, the learning rate for weights is divided by 10 at $1/2$ and $3/4$ of the maximum number of epochs. The weight parameters are initialized by He's initialization \cite{He2015}. We used the cross-entropy loss of $l(z, W, M)$. The experimental setting of the proposed method is based on \cite{Shirakawa2018}; the sample size is $\lambda = 2$, the learning rate is $\eta_{\theta} = 1 / d$, and the initial distribution parameters is $\theta^{(0)} = 0.5$.

%%%%%%%%%%%%%%%%%%%%%%%%%
\subsection{Unit Selection of the Fully-Connected Neural Network}
\subsubsection{Experimental Setting}
In this experiment, we use a fully-connected neural network with three hidden layers of 784 units as the base structure, and we select the units in each hidden layer. The MNIST dataset, which is a 10 class handwritten digits dataset consisting of 60,000 training and 10,000 test data of $28 \times 28$ gray-scaled images, is used. We use the pixel values as the inputs and determine the existence of the units in hidden layers according to the binary vector $M$. The $i$-th unit is active if $m_i = 1$ and inactive if $m_i = 0$. The output of the $i$-th unit is represented by $m_{i} F(X_{i})$, where $F$ and $X_i$ denote the activation function and the input for the activation of the $i$-th unit, respectively. We use rectified linear unit (ReLU) and softmax activation as $F$ in hidden layers and in an output layer, respectively. When $m_i = 0$, the connections to/from the $i$-th unit can be omitted; that is, the active number of weights decreases. The dimension of $\theta$, the total number of hidden units, is $d = 2352$. This task is simple, but we can check how the proposed penalty term works.

We set the mini-batch size to $\bar{N} = 32$ and the number of training epochs to 500 in the proposed method, while the mini-batch size to $\bar{N} = 64$ and the number of training epochs is set to 1000 in other methods. Under these settings, the number of data samples used in one iteration and the number of iterations for parameters update become identical in all methods, where the number of iterations is about $9.5 \times 10^{5}$. We initialize the learning rate for $W$ by 0.01. In this experiment, we change the coefficient $\epsilon'$ as $2^{-6}$, $2^{-7}$, $2^{-8}$, $2^{-9}$, $0$, $-2^{-3}$, and $-2^{0}$ to check the effect of the penalty term.\footnote[1]{The negative value of $\epsilon'$ encourages the increase of the number of active units.} Since each bit decides whether or not its corresponding unit is active, we simply use the same coefficients of model complexity for each unit, $c = (1, \dots, 1)^{\T}$.

To evaluate the proposed method's performance, we report the experimental result of the fixed neural network structures with the various numbers of units. We manually and uniformly remove the units in the hidden layers and control the number of weights. As the network structures are stochastically sampled in the training phase of the dynamic structure optimization, our method is somewhat similar to that of stochastic network models, such as Dropout \cite{Srivastava2014}. We also report the result using Dropout with a dropout rate of 0.5 for comparison. Note that the aim of a dropout is to prevent overfitting; thus, all units are kept in the test phase.

\renewcommand{\subfigcapmargin}{15pt} % space parameter between figures.
\begin{figure}[tbp]
    \begin{center}
    \subfigure[{\small Unit selection}]{
        \centering
        \includegraphics[width=0.45\linewidth]{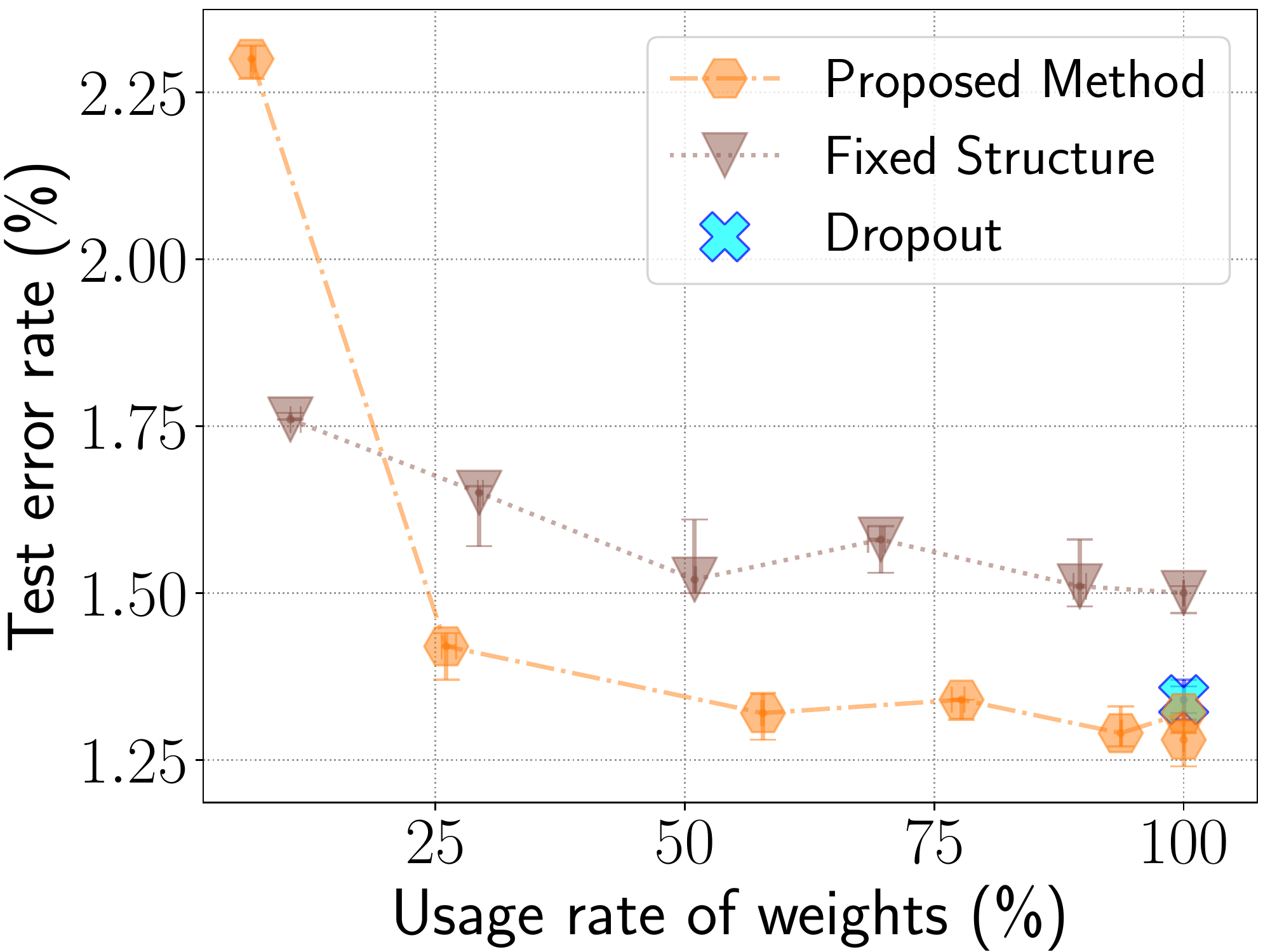}
        \label{fig::unit_sel_error}
    }
    \subfigure[{\small Connection selection}]{
        \centering
        \includegraphics[width=0.45\linewidth]{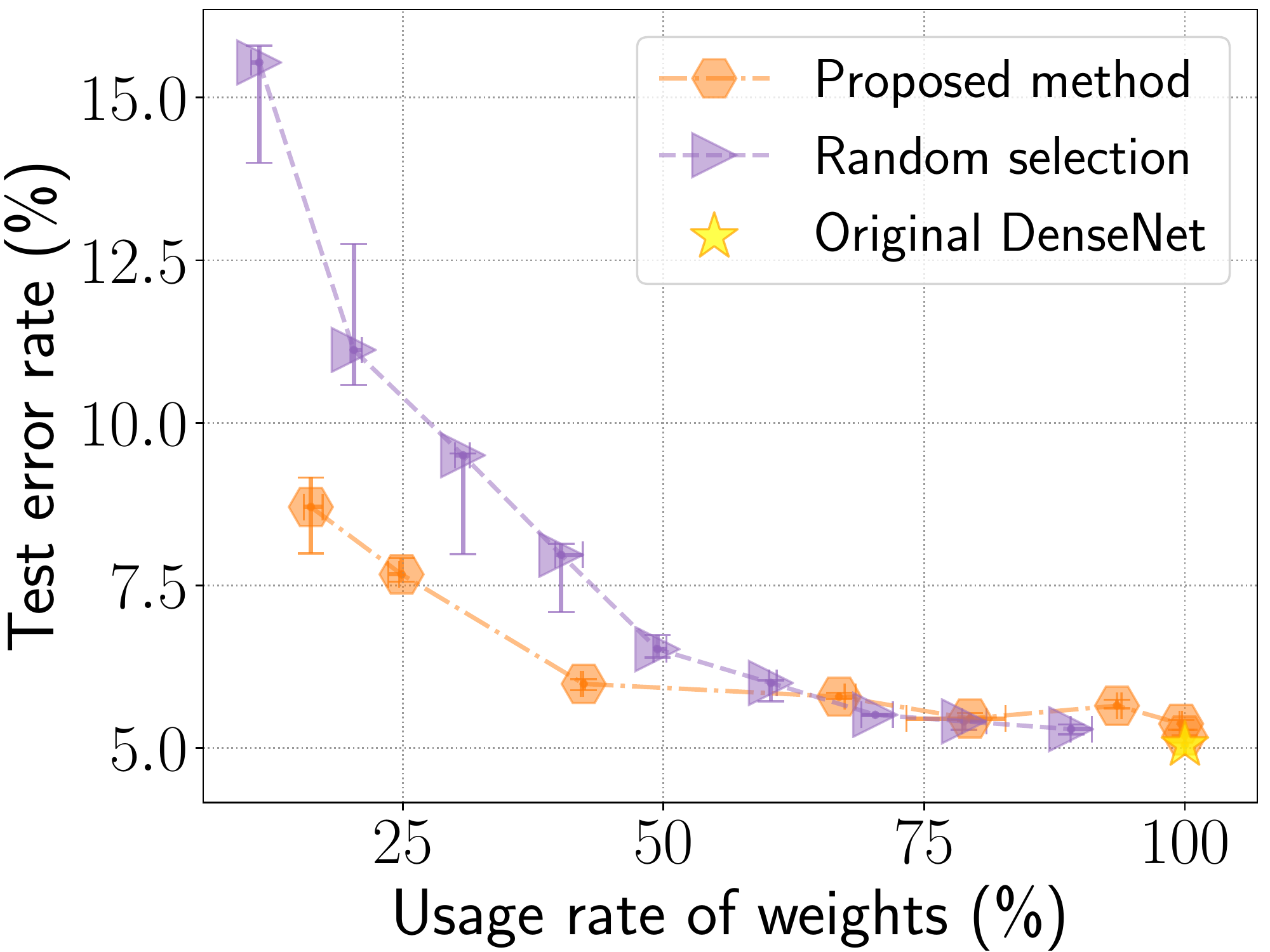}
        \label{fig::connect_sel_error}
    }
    \caption{The relationship between the weight usage rate and test error rates of (a) unit selection of the fully-connected neural network and (b) the connection selection of DenseNet. The median values and 25\% and 75\% quantile values of each over five independent trials are plotted.}
    \end{center}
    \label{fig:err}
\end{figure}

\subsubsection{Result and Discussion}
Figure \ref{fig::unit_sel_error} shows the relation between the weight usage rate and test error rates of the proposed method, the fixed structure, and Dropout. The median values and the 25\% and 75\% quantile values of each over five independent trials are plotted.

Comparing the proposed method and fixed structure, the proposed method outperforms the fixed structure over the 25\% usage rate of weights. In the fixed structure, the error rate gradually increases as the usage rate of weights decreases. The performance of the proposed method deteriorates when its weights usage rate is approximately 6\%. This indicates that the proposed method can control the usage rate of weights by changing the penalty coefficient of $\epsilon'$ and remove the units while still maintaining its performance. The structures obtained by the different $\epsilon'$ settings create a trade-off curve between the model complexity and performance.

Comparing the proposed method and the original structure (i.e., the fixed structure with the $100\%$ weight usage rate), the proposed method outperforms the original structure in the usage rate of $25\%$ to $100\%$. Remarkably, when $\epsilon' < 0$, although all units are selected after the training procedure (i.e., the structure is the same as the original structure), the performance improves. Additionally, dropout training also improves performance. Based on these results, stochastic training appears to improve prediction performance. Dropout, however, cannot control the weight usage rate, but the proposed method can reduce the number of used weights without significant performance deterioration.

Table \ref{tb::unit_sel} shows a summary of median values of the number of selected units in each layer. We observe that the proposed method preferentially removes units in the second and third hidden layers. Therefore, the proposed method removes the units selectively rather than at random.

The computational time for training by the proposed method is almost the same as that required by the fixed structure. Even if we run several different penalty coefficient $\epsilon'$ settings to obtain additional trade-off structures, the total computational time of the structure search more or less increases several times over. This is reasonable more than the hyperparameter optimization-based structure optimization.

\begin{table}[bt]
{\small
\begin{minipage}[c]{.47\textwidth}
	\begin{center}
		\caption{The numbers of selected units in each hidden layer in the unit selection experiment.
		}
        \label{tb::unit_sel}
		\begin{tabular}{crrr}
		\toprule
		Weight & $1$st & $2$nd & $3$rd \\
		usage rate & \multicolumn{3}{c}{layer} \\
		\midrule
		6.6\% ($\epsilon' = 2^{-6}$) & 92 & 31 & 24 \\
		26.1\% ($\epsilon' = 2^{-7}$) & 340 & 141 & 126 \\
		57.8\% ($\epsilon' = 2^{-8}$) & 599 & 380 & 379 \\
		77.7\% ($\epsilon' = 2^{-9}$) & 704 & 544 & 567 \\
		93.7\% ($\epsilon' = 0$) & 770 & 717 & 724 \\
		100\% ($\epsilon' = -2^{-3}$) & 784 & 784 & 784 \\
		\bottomrule
		\end{tabular}
	\end{center}
\end{minipage}	
\hfill%
\begin{minipage}[c]{.47\textwidth}
	\begin{center}
		\caption{The numbers of selected connections in each block in the connection selection experiment.}
        \label{tb::connect_sel}
		\begin{tabular}{crrr}
		\toprule
		Weight & $1$st & $2$nd & $3$rd \\
		usage rate & \multicolumn{3}{c}{block} \\
		\midrule
		15.8\% ($\epsilon' = 2^{-2}$) & 36 & 6 & 20 \\
		24.1\% ($\epsilon' = 2^{-3}$) & 45 & 8 & 49 \\
		42.3\% ($\epsilon' = 2^{-4}$) & 57 & 39 & 67 \\
		67.5\% ($\epsilon' = 2^{-5}$) & 59 & 61 & 76 \\
		80.3\% ($\epsilon' = 2^{-6}$) & 65 & 64 & 80 \\
		100\% ($\epsilon' = -2^{0}$) & 91 & 91 & 91 \\
		\bottomrule
		\end{tabular}
	\end{center}
\end{minipage}
}
\end{table}

%%%%%%%%%%%%%%%%%%%%%%%%%%%
\subsection{Connection Selection of DenseNet}
\subsubsection{Experimental Setting}
We use DenseNet \cite{Huang2017} as the base network structure; it is composed of several dense blocks and transition layers. The dense block consists of $L_{\mathrm{block}}$ layers, each of which implements a non-linear transformation with batch normalization (BN) \cite{Ioffe2015} followed by the ReLU activation and the $3 \times 3$ convolution. In the dense block, the $l$-th layer receives the outputs of all the preceding layers as inputs that are concatenated on the channel dimension. The size of the output feature-maps in the dense block is the same as that of the input feature-maps. The transition layer is located between the dense blocks and consists of the batch normalization, ReLU activation, and the $1 \times 1$ convolution layer, which is followed by $2 \times 2$ average pooling. The detailed structure of DenseNet can be found in \cite{Huang2017}. Unlike \cite{Huang2017}, however, we do not use Dropout.

We optimize the connections in the dense blocks using the CIFAR-10 dataset, which contains 50,000 training and 10,000 test data of $32 \times 32$ color images in the 10 different classes. During the preprocessing and data augmentation, we use the standardization, padding, and cropping for each channel, and this is  followed by randomly horizontal flipping. The setting details are the same as in \cite{Shirakawa2018}.

We determine the existence of the connections between the layers in each dense block according to the binary vector $M$. As was done in \cite{Shirakawa2018}, we use a simple DenseNet structure with a depth of 40 that contains three dense blocks with $L_{\mathrm{block}} = 12$ and two transition layers. In this setup, the dimension of $M$ and $\theta$ becomes $d = 273$. We vary the coefficient $\epsilon'$ as $2^{-2}$, $2^{-3}$, $2^{-4}$, $2^{-5}$, $2^{-6}$, $0$, $-2^{-3}$, and $-2^{0}$ to assess the effect of the penalty term. Additionally, we set the coefficients of the model complexity $c_i$ to match the number of weights corresponding to the $i$-th connection.

For the proposed method, we set the mini-batch size to $\bar{N} = 32$ and the number of training epochs to 300. For the other methods, the mini-batch size is set to $\bar{N} = 64$ and the number of training epochs to 600. With these settings, the number of iterations for parameter updates become identical in all methods, where the number of iterations is about $4.7 \times 10^{5}$. We initialize the learning rate for $W$ by 0.1.

We also report the result when the connections are removed randomly. We repeatedly sample the binary vector $M$ such that the weight usage rate becomes the target percentage, and then we train the fixed network.

\subsubsection{Result and Discussion}
Figure \ref{fig::connect_sel_error} shows the relations between the weight usage rate and test error rate. Comparing the proposed method and random selection, the proposed method outperforms the random selection in the 15\% to 40\% weight usage rate. In the random selection method, important connections might be lost when the usage rate of weights is less than 40\%. In contrast, the proposed method can selectively remove the number of weights without increasing the test errors, so it does not eliminate important connections. However, the difference between the test error rates of the random selection and the proposed method is not significant when the weight usage rate exceeds 60\%. This result indicates that a small number of connections in DenseNet can be randomly removed without performance deterioration, meaning that DenseNet might be redundant; the proposed method can moderate increase of the test error rate within 1\% in the 40\% weight usage rate.

Table \ref{tb::connect_sel} summarizes the median values of the number of selected connections in each block.
The proposed method preferentially remove the connections in the first and second blocks when $\epsilon' = 0$ to $2^{-5}$, but these deletions do not have a significant impact on performance. When $\epsilon' = 2^{-4}$ to $2^{-2}$, the proposed method actively removes the connections in the second block, so the obtained structure can reduce the performance deterioration more than random selection.

\begin{figure*}[tbp]
    \centering
    \includegraphics[width=0.99\linewidth]{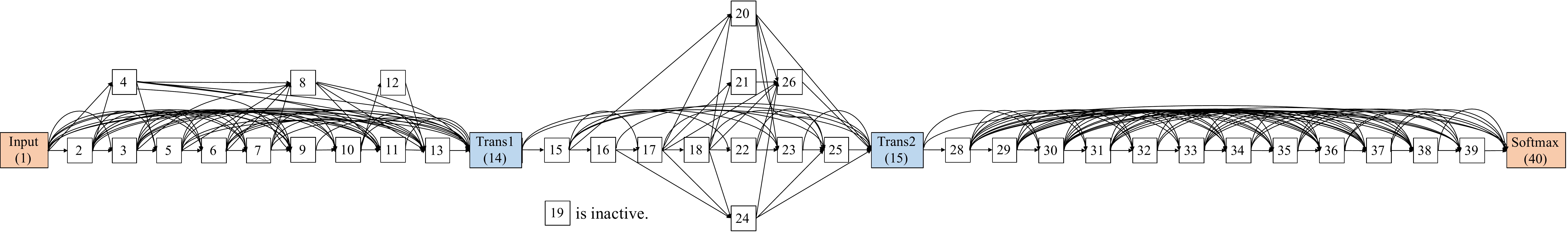}
    \caption{The obtained DenseNet structure in the case of $\epsilon' = 2^{-4}$ on a typical single run. The numbers in the cells represent the depth of each layer in the original DenseNet structure. Cells placed in the same column locate the same depth, and the depth of this DenseNet structure is 32.}
    \label{fig::structure}
\end{figure*}

Figure \ref{fig::structure} shows the structure obtained by the proposed method when $\epsilon' = 2^{-4}$ on a typical single run. As we can see, the second block in this structure, which is between `Trans1' and `Trans2' cells, becames sparser and wider than the first and third blocks. Interestingly, the second block became a wide structure through removing the connections between its layers. This result might suggest that wide structures may be able to improve performance with limited computing resources. Several works, such as \cite{Zagoruyko2016}, report that widening layers improves the predictive performance; our findings may also support these wide networks. We would like to emphasize that it is not easy to manually design a structure, such as that shown in Figure \ref{fig::structure}, due to the differing connectivities in each block. 

Finally, we note that the amount of computational time required by our structure for training is not  significantly greater than that required by random selection, meaning that our proposed structure optimization is computationally efficient.

\section{Conclusion}
\label{sec::conclusion}
In this paper, we propose a method  of controlling model complexity by adding a penalty term to the objective function involved in the dynamic structure optimization of DNNs. We incorporate a penalty term dependent on structure parameters into the loss function and consider its expectation under the multivariate Bernoulli distribution to be the objective function. We derive a modified update rule that enables us to control model complexity.

In the experiment on unit selection, the proposed method outperforms the fixed structure in terms of a 25 to 100\% weight usage rate. In the connection selection experiment, the proposed method also outperforms random selection in the small number of weights and preferentially removing insignificant connections during the training. Upon checking the obtained structure, it is found that the intermediate block became a wide structure. 

As the increased amount of the computational time required by the proposed method is not significant, we can take the trade-off between model complexity and performance with an acceptable computational cost. Our method requires training only once, whereas the pruning methods, such as that in \cite{Liu2017}, require the retraining after pruning. 

In future work, we will apply the proposed method to the architecture search method for more complex neural network structures, such as that proposed in \cite{Akimoto2019}. Additionally, we should evaluate the proposed penalty term using different datasets. Another possible future work is modifying the proposed method so that it can use other types of the model complexity criteria, such as FLOPs of neural networks.

\section*{Acknowledgment}
This work is partially supported by the SECOM Science and Technology Foundation.

\bibliographystyle{splncs04}
\bibliography{body/reference}

\begin{thebibliography}{10}
\providecommand{\url}[1]{\texttt{#1}}
\providecommand{\urlprefix}{URL }
\providecommand{\doi}[1]{https://doi.org/#1}

\bibitem{Akimoto2019}
Akimoto, Y., Shirakawa, S., Yoshinari, N., Uchida, K., Saito, S., Nishida, K.:
  {Adaptive Stochastic Natural Gradient Method for One-Shot Neural Architecture
  Search}. In: International Conference on Machine Learning (ICML). pp.
  171--180 (2019)

\bibitem{Amari1998}
Amari, S.: {Natural Gradient Works Efficiently in Learning}. Neural Computation
   \textbf{10}(2),  251--276 (1998). \doi{10.1162/089976698300017746}

\bibitem{Dong2018}
Dong, J., Cheng, A., Juan, D., Wei, W., Sun, M.: {PPP-Net: Platform-aware
  Progressive Search for Pareto-optimal Neural Architectures}. In:
  International Conference on Learning Representations (ICLR) Workshop (2018)

\bibitem{Elsken2018}
Elsken, T., Metzen, J.H., Hutter, F.: {Efficient Multi-objective Neural
  Architecture Search via Lamarckian Evolution}. In: International Conference
  on Learning Representations (ICLR) (2019)

\bibitem{Han2015}
Han, S., Pool, J., Tran, J., Dally, W.J.: {Learning both Weights and
  Connections for Efficient Neural Networks}. In: Neural Information Processing
  Systems (NIPS) (2015)

\bibitem{He2015}
He, K., Zhang, X., Ren, S., Sun, J.: {Delving Deep into Rectifiers: Surpassing
  Human-Level Performance on ImageNet Classification}. In: IEEE International
  Conference on Computer Vision (ICCV). pp. 1026--1034 (2015).
  \doi{10.1109/ICCV.2015.123}

\bibitem{Huang2017}
Huang, G., Liu, Z., van~der Maaten, L., Weinberger, K.Q.: {Densely Connected
  Convolutional Networks}. In: IEEE Conference on Computer Vision and Pattern
  Recognition (CVPR). pp. 2261--2269 (2017). \doi{10.1109/CVPR.2017.243}

\bibitem{Ioffe2015}
Ioffe, S., Szegedy, C.: {Batch Normalization: Accelerating Deep Network
  Training by Reducing Internal Covariate Shift}. In: International Conference
  on Machine Learning (ICML). pp. 448--456 (2015)

\bibitem{Kandasamy2018}
Kandasamy, K., Neiswanger, W., Schneider, J., Poczos, B., Xing, E.: {Neural
  Architecture Search with Bayesian Optimisation and Optimal Transport}. In:
  Neural Information Processing Systems (NIPS) (2018)

\bibitem{Liu2019}
Liu, H., Simonyan, K., Yang, Y.: {DARTS: Differentiable Architecture Search}.
  In: International Conference on Learning Representations (ICLR) (2019)

\bibitem{Liu2017}
Liu, Z., Li, J., Shen, Z., Huang, G., Yan, S., Zhang, C.: {Learning Efficient
  Convolutional Networks through Network Slimming}. In: IEEE International
  Conference on Computer Vision (ICCV). pp. 2736--2744 (2017)

\bibitem{Okuta2017}
Okuta, R., Unno, Y., Nishino, D., Hido, S., Loomis, C.: {CuPy: A
  NumPy-Compatible Library for NVIDIA GPU Calculations}. In: Workshop on
  Machine Learning Systems (LearningSys) in The 31st Annual Conference on
  Neural Information Processing Systems (NIPS) (2017)

\bibitem{Ollivier2017}
Ollivier, Y., Arnold, L., Auger, A., Hansen, N.: {Information-Geometric
  Optimization Algorithms: A Unifying Picture via Invariance Principles}.
  Journal of Machine Learning Research  \textbf{18}(18),  1--65 (2017)

\bibitem{Pham2018}
Pham, H., Guan, M., Zoph, B., Le, Q.V., Dean, J.: {Efficient Neural
  Architecture Search via Parameters Sharing}. In: International Conference on
  Machine Learning (ICML). pp. 4095--4104 (2018)

\bibitem{Real2017}
Real, E., Moore, S., Selle, A., Saxena, S., Suematsu, Y.L., Tan, J., Le, Q.,
  Kurakin, A.: {Large-Scale Evolution of Image Classifiers}. In: International
  Conference on Machine Learning (ICML). pp. 2902--2911 (2017)

\bibitem{Saito2018}
Saito, S., Shirakawa, S., Akimoto, Y.: {Embedded Feature Selection Using
  Probabilistic Model-Based Optimization}. In: Genetic and Evolutionary
  Computation Conference (GECCO) Companion. pp. 1922--1925 (2018).
  \doi{10.1145/3205651.3208227}

\bibitem{Shirakawa2018}
Shirakawa, S., Iwata, Y., Akimoto, Y.: {Dynamic Optimization of Neural Network
  Structures Using Probabilistic Modeling}. In: The 32nd AAAI Conference on
  Artificial Intelligence (AAAI-18). pp. 4074--4082 (2018)

\bibitem{Srivastava2014}
Srivastava, N., Hinton, G., Krizhevsky, A., Sutskever, I., Salakhutdinov, R.:
  {Dropout: A Simple Way to Prevent Neural Networks from Overfitting}. Journal
  of Machine Learning Research  \textbf{15},  1929--1958 (2014)

\bibitem{Suganuma2017}
Suganuma, M., Shirakawa, S., Nagao, T.: {A Genetic Programming Approach to
  Designing Convolutional Neural Network Architectures}. In: Genetic and
  Evolutionary Computation Conference (GECCO). pp. 497--504 (2017).
  \doi{10.1145/3071178.3071229}

\bibitem{Tan2018}
Tan, M., Chen, B., Pang, R., Vasudevan, V., Sandler, M., Howard, A., Le, Q.V.:
  {MnasNet: Platform-Aware Neural Architecture Search for Mobile}. In: The IEEE
  Conference on Computer Vision and Pattern Recognition (CVPR) (June 2019)

\bibitem{Tokui2015}
Tokui, S., Oono, K., Hido, S., Clayton, J.: {Chainer: a Next-Generation Open
  Source Framework for Deep Learning}. In: Workshop on Machine Learning Systems
  (LearningSys) in Neural Information Processing Systems (NIPS). pp.~1--6
  (2015)

\bibitem{Zagoruyko2016}
Zagoruyko, S., Komodakis, N.: {Wide Residual Networks}. In: British Machine
  Vision Conference (BMVC). pp. 87.1--87.12 (2016). \doi{10.5244/C.30.87}

\bibitem{Zoph2017}
Zoph, B., Le, Q.V.: {Neural Architecture Search with Reinforcement Learning}.
  In: International Conference on Learning Representations (ICLR) (2017)

\end{thebibliography}

\end{document}